\documentclass[10pt,twocolumn,letterpaper]{article}

\usepackage{cvpr}
\usepackage{times}
\usepackage{epsfig}
\usepackage{graphicx}
\usepackage{amsmath}
\usepackage{amssymb}
\usepackage{bm}
\usepackage{enumitem}
\usepackage{mathtools}
\usepackage{multirow}
\usepackage[htt]{hyphenat}
\usepackage{fontawesome}
\usepackage{pifont}

\usepackage[pagebackref=true,breaklinks=true,letterpaper=true,colorlinks,bookmarks=false]{hyperref}

\DeclareMathOperator*{\argmax}{argmax}
\DeclareMathOperator*{\argmin}{argmin}

\cvprfinalcopy 

\def\cvprPaperID{5633} 

\ifcvprfinal\fi
\begin{document}

\title{ClusterVO: Clustering Moving Instances and Estimating Visual Odometry \\ for Self and Surroundings}

\author{Jiahui Huang$^1$ \quad Sheng Yang$^{2}$ \quad Tai-Jiang Mu$^1$ \quad Shi-Min Hu$^1$\thanks{corresponding author.}\\
$^1$BNRist, Department of Computer Science and Technology, Tsinghua University,
Beijing \\ $^2$Alibaba Inc., China \\
{\tt\small huang-jh18@mails.tsinghua.edu.cn, shengyang93fs@gmail.com} \\
{\tt\small taijiang@tsinghua.edu.cn, shimin@tsinghua.edu.cn} \\
{\small\url{https://youtu.be/paK-WCQpX-Y}}
}

\maketitle

\begin{abstract}
   We present ClusterVO, a stereo Visual Odometry which simultaneously clusters and estimates the motion of both ego and surrounding rigid clusters/objects.
   Unlike previous solutions relying on batch input or
   imposing priors on scene structure or dynamic object models, 
   ClusterVO is online, general and thus can be used in various scenarios including indoor scene understanding and autonomous driving.
   At the core of our system lies a multi-level probabilistic association mechanism and a heterogeneous Conditional Random Field (CRF) clustering approach combining semantic, spatial and motion information to jointly infer cluster segmentations online for every frame. 
   The poses of camera and dynamic objects are instantly solved through a sliding-window optimization.
   Our system is evaluated on Oxford Multimotion and KITTI dataset both quantitatively and qualitatively, reaching comparable results to state-of-the-art solutions on both odometry and dynamic trajectory recovery.
\end{abstract}

\section{Introduction}

Understanding surrounding dynamic objects is an important step beyond ego-motion estimation in the current visual Simultaneous Localization and Mapping (SLAM) community for the frontier requirements of advanced Augmented Reality (AR) or autonomous things navigation:
In typical use cases of Dynamic AR, these dynamics need to be explicitly tracked to enable interactions of virtual object with moving instances in the real world.
In outdoor autonomous driving scenes, a car should not only accurately localize itself but also reliably sense other moving cars to avoid possible collisions.

Despite the above need from emerging applications to perceive scene motions, most classical SLAM systems~\cite{bescos2018dynaslam,Keller20133DV,kim2016effective,mur2017orb} merely regard dynamics as outliers during pose estimation.
Recently, advances in vision and robotics have demonstrated us with new possibilities of developing motion-aware Dynamic SLAM systems by coupling various different vision techniques like detection and tracking~\cite{bhat2018unveiling,redmon2017yolo9000,sharma2018beyond}. Nevertheless, currently these systems are often tailored for special use cases: For indoor scenes where dense RGB{-}D data are available, geometric features including convexity or structure regularities are used to assist segmentation~\cite{caccamo2017joint,ruenz2017icra,runz2018maskfusion,strecke2019fusion,xu2019mid}. For outdoor scenes, object priors like car sizes or road planar structure are exploited to constrain the solution spaces~\cite{barsan2018robust,li2018stereo,luiten2020track,yang2019cubeslam}. These different assumptions render existing algorithms hardly applicable to general dynamic scenarios.
Contrarily, ClusterSLAM~\cite{huang2019clusterslam} incorporates no scene prior, but it acts as a backend instead of a full system whose performance relies heavily on the landmark tracking and association quality.

\begin{figure}[t]
    \centering
    \includegraphics[width=\linewidth]{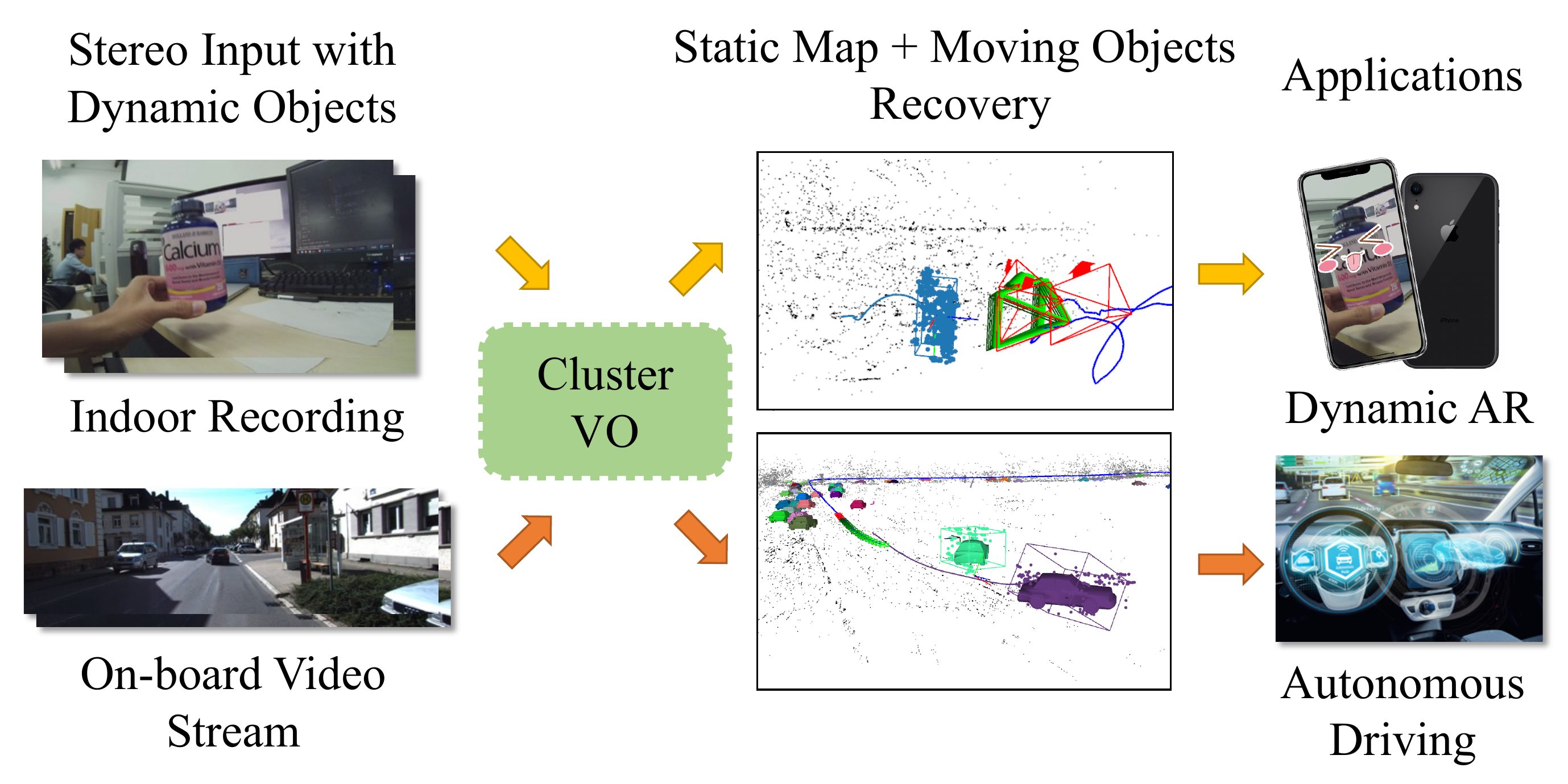}
    \caption{Our proposed system \emph{ClusterVO} can simultaneously recover the camera ego-motion as well as cluster trajectories.}
    \label{fig:teaser}
\end{figure}

To bridge the above gap in current Dynamic SLAM solutions in the literature,
we propose ClusterVO, a stereo visual odometry system for dynamic scenes, which simultaneously optimizes 
the poses of camera and multiple moving objects, regarded as clusters of point landmarks, in a unified manner, achieving a competitive frame-rate with promising tracking and segmentation ability as listed in Table~\ref{tbl:comp}.
Because no geometric or shape priors on the scene or dynamic objects are imposed, our proposed system is general and adapts to many various applications ranging from autonomous driving, indoor scene perception to augmented reality development.
Our novel strategy is solely based on sparse landmarks and 2D detections~\cite{redmon2017yolo9000}; to make use of such a lightweight representation, we propose a robust multi-level probabilistic association technique to efficiently track both low-level features and high-level detections over time in the 3D space.
Then a highly-efficient heterogeneous CRF jointly considering semantic bounding boxes, spatial affinity and motion consistency is applied to discover new clusters, cluster novel landmarks and refine existing clusterings.
Finally, Both static and dynamic parts of the scene are solved in a sliding-window optimization fashion.

\section{Related Works}

\begin{table}[t]
\centering
\setlength{\tabcolsep}{5.5pt}
\caption{Comparison with other dynamic SLAM solutions. \faCamera: Sensor(s) used. \faHome: Applicable in indoor scene? \faAutomobile: Applicable in outdoor driving scenarios? \faCubes: Recover poses of moving rigid bodies? \faSpaceShuttle: Is online? `NR' represents single Non-Rigid body.}
\label{tbl:comp}
\vspace{0.3em}
\small
\begin{tabular}{l|cccccc}
\hline
                   &  \faCamera   & \faHome   & \faAutomobile   & \faCubes & \faSpaceShuttle    & FPS \\ \hline
ORB-SLAM2~\cite{mur2017orb}           & Multiple & \checkmark & \checkmark &            & \checkmark & 10  \\
DynamicFusion~\cite{newcombe2015dynamicfusion}      & RGB-D    & \checkmark &            & NR  & \checkmark & -   \\
MaskFusion~\cite{runz2018maskfusion}         & RGB-D    & \checkmark &            & \checkmark & \checkmark & 30  \\
Li \etal~\cite{li2018stereo}           & Stereo   &            & \checkmark & \checkmark &            & 5.8 \\
DynSLAM~\cite{barsan2018robust}           & Stereo   &            & \checkmark & \checkmark & \checkmark & 2   \\
ClusterSLAM~\cite{huang2019clusterslam}        & Stereo   & \checkmark & \checkmark & \checkmark &            & 7   \\ \hline
\textbf{ClusterVO} & Stereo   & \checkmark & \checkmark & \checkmark & \checkmark & \textbf{8} \\ \hline
\end{tabular}
\end{table}

\noindent\textbf{Dynamic SLAM / Visual Odometry.}
Traditional SLAM or VO systems are based on static scene assumption and dynamic contents need to be carefully handled which would otherwise lead to severe pose drift.
To this end, some systems explicitly detect motions and filter them either with motion consistency~\cite{dai2018rgb,Keller20133DV,kim2016effective} or object detection modules~\cite{bescos2018dynaslam,yu2018ds,zhong2018detect}.
The idea of simultaneously estimating ego motion and multiple moving rigid objects, same as our formulation, originated from the seminal SLAMMOT~\cite{wang2007simultaneous} project. 
Follow-ups like \cite{caccamo2017joint,ruenz2017icra,runz2018maskfusion,strecke2019fusion,xu2019mid} use RGB-D as input and reconstruct dense models for the indoor scene along with moving objects. For better segmentation of object identities, \cite{runz2018maskfusion,xu2019mid} combine heavy instance segmentation module and geometric features.
\cite{kumar2016spatiotemporal,tagliasacchi2015robust,tzionas2016reconstructing} can track and reconstruct rigid object parts on a predefined articulation template (\eg human hands or kinematic structures).
\cite{eckenhoff2019tightly,qiu2019tracking} couple existing visual-inertial system with moving objects tracked using markers.
Many other methods are specially designed for road scenes by exploiting modern vision modules~\cite{behl2017bounding,li2018stereo,luiten2020track,menze2015object,nair2020multi,xiang2015robust}.
Among them, \cite{li2018stereo} proposes a batch optimization to accurately track the motions of moving vehicles but a real-time solution is not presented.

Different from ClusterSLAM~\cite{huang2019clusterslam}, which is based on motion affinity matrices for hierarchical clustering and SLAM, this work focuses on developing a relatively light-weight visual odometry, and faces challenges from real-time clustering and state estimation.

\noindent\textbf{Object Detection and Pose Estimation.}
With the recent advances in deep learning technologies, the performance of 2D object detection and tracking have been boosted~\cite{bhat2018unveiling,girshick2015fast, he2017mask,liu2016ssd,redmon2017yolo9000,sharma2018beyond}.
Detection and tracking in 3D space from video sequences is a relatively unexplored area due to the difficulty in the 6-DoF~(six degrees of freedom) pose estimation. In order to accurately estimate 3D positions and poses, many methods~\cite{grabner2019gp2c,li2019stereo} leverages a predefined object template or priors to jointly infer object depth and rotations.
In ClusterVO, the combination of low-level geometric feature descriptors and semantic detections inferred simultaneously in the localization and mapping process can provide additional cues for efficient tracking and accurate object pose estimation.

\section{ClusterVO}

\begin{figure*}
\begin{center}
\includegraphics[width=\linewidth]{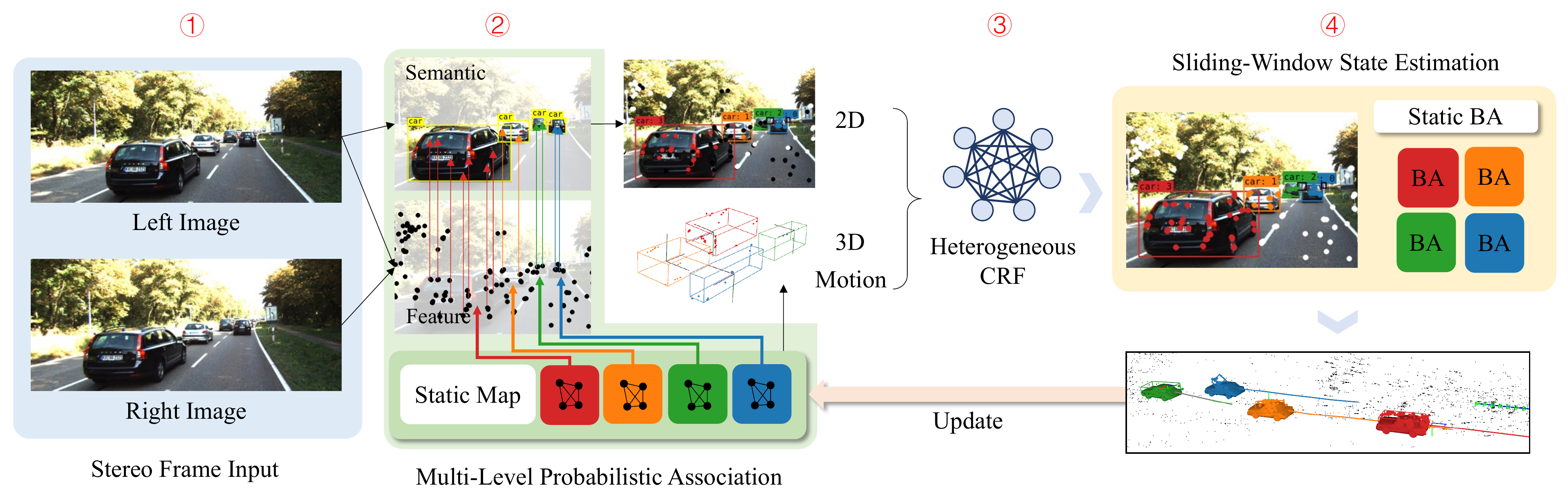}
\end{center}
   \caption{\textbf{Pipeline of ClusterVO.} \ding{172} For each incoming stereo frame ORB features and semantic bounding boxes are extracted. \ding{173} We apply multi-level probabilistic association to associate features with landmarks and bounding boxes with existing clusters. \ding{174} Then we cluster the landmarks observed in the current frame into different rigid bodies using the Heterogeneous CRF module. \ding{175} The state-estimation is performed in a sliding window manner with specially designed keyframe mechanism. Optimized states are used to update the static maps and clusters.}
\label{fig:pipeline}
\end{figure*}

ClusterVO takes synchronized and calibrated stereo images as input, and outputs camera and object pose for each frame.
For each incoming frame, semantic bounding boxes are detected using YOLO object detection network~\cite{redmon2017yolo9000}, and ORB features~\cite{rublee2011orb} are extracted and matched across stereo images.
We first associate detected bounding boxes and extracted features to previously found clusters and landmarks, respectively, through a multi-level probabilistic association formulation (Sec. \ref{subsec:assoc}).
Then, we perform heterogeneous conditional random field (CRF) over all features with associated map landmarks to determine the cluster segmentation for current frame (Sec. \ref{subsec:crf}).
Finally, the state estimation step optimizes all the states
over a sliding window with marginalization and a smooth motion prior (Sec. \ref{subsec:slam}).
The pipeline is illustrated in Figure~\ref{fig:pipeline}.

\noindent\textbf{Notations.}
At frame $t$, ClusterVO outputs: the pose of the camera $\mathbf{P}_t^{\mathbf{c}}$
in the global reference frame,
the state of
all clusters (rigid bodies) $\{\bm{x}_t^{\mathbf{q}}\}_{\mathbf{q}}$, and the state of all landmarks $\bm{x}_t^{\mathbf{L}}$.
The $\mathbf{q}$-th cluster state $\bm{x}_t^{\mathbf{q}} = ( \mathbf{P}_t^{\mathbf{q}}, \bm{v}_t^{\mathbf{q}} ) $ contains current 6-DoF pose $\mathbf{P}_t^{\mathbf{q}} \in \mathrm{SE(3)}$ and current linear speed in 3D space $\bm{v}_t^{\mathbf{q}} \in \mathbb{R}^3$.
Specially we use $\mathbf{q}=0$ to denote the static scene for convenience. 
Hence $\forall t, \mathbf{P}_t^{0} \equiv \mathbf{I}, \bm{v}_t^{0} \equiv \mathbf{0}$.
As a short hand, we denote the transformation from coordinate frame $\mathbf{a}$ to frame $\mathbf{b}$
as $\mathbf{T}_t^{\mathbf{ab}} \coloneqq (\mathbf{P}_t^\mathbf{a})^{-1} \mathbf{P}_t^\mathbf{b}$.
For the landmark state $\bm{x}_t^{\mathbf{L}} = \{ (\bm{p}_t^i, \mathbf{q}^i, w^i) \}_{i}$, each landmark $i$ has the property of its global position $\bm{p}_t^i \in \mathbb{R}^3$, the cluster assignment $\mathbf{q}^i$ and its confidence $w^i \in \mathbb{N}^{+}$ defining the cluster assignment confidence.
For observations, we denote the location of the $k$-th low-level ORB stereo feature extracted
at frame $t$ as $\bm{z}_t^k = (u_L, v_L, u_R) \in \mathbb{R}^3$, and the $m$-th high-level semantic bounding box detected at frame $t$ as $\bm{B}_t^m$.
Assuming the feature observation $\bm{z}_t^k$ is subject to a Gaussian noise with covariance $_{\bm{z}}\Sigma$, the noise of the triangulated points $\bm{Z}_t^i$ in camera space can be calculated as $_{\bm{Z}}\Sigma_t^i = \mathbf{J}_{\pi^{-1}} (\prescript{}{\bm{z}}{\Sigma}) \mathbf{J}_{\pi^{-1}}^\top$, where $\pi$ is the stereo projection function, $\pi^{-1}$ is the corresponding back-projection function and $\mathbf{J}_{f}$ is the Jacobian matrix of function $f$.

For generality, we do not introduce a category-specific canonical frame for each cluster.
Instead, we initialize the cluster pose $\mathbf{P}_t^{\mathbf{q}}$ with the center and the three principal orthogonal directions of the landmark point clouds belonging to the cluster as the translational and rotational part respectively and track the relative pose ever since.

\subsection{Multi-level Probabilistic Association}
\label{subsec:assoc}

For the landmarks on static map (\ie $\mathbf{q}^i = 0$), the features can be robustly associated by nearest neighbour search and descriptor matching~\cite{mur2017orb}. 
However, tracking dynamic landmarks which move fast on the image space is not a trivial task.
Moreover, we need to associate each detected bounding box $\bm{B}_t^m$ to an existing map cluster if possible, 
which is required in the succeeding Heterogeneous CRF module.

To this end, we propose a multi-level probabilistic association scheme for dynamic landmarks (\ie $\mathbf{q}^i \neq 0$), assigning low-level feature observation $\bm{z}_t^k$ to its source landmark id $k\rightarrow i$ and high-level bounding box $\bm{B}_t^m$ to a cluster $m\rightarrow \mathbf{q}$.
The essence of the probabilistic approach is to model the position of a landmark by a Gaussian distribution with mean $\bm{p}_t^i$ and covariance $\Sigma_t^i$ and consider the uncertainty throughout the matching.

Ideally, $\Sigma_t^i$ should be extracted from the system information matrix from the last state estimation step, but the computation burden is heavy. We hence approximate $\Sigma_t^i$ as transformed $\prescript{}{\bm{Z}}{\Sigma^i_{t' < t}}$ with the smallest determinant,  \ie:
\begin{equation}
    \Sigma_t^i \coloneqq \mathbf{R}_{t'}^\mathbf{c} \prescript{}{\bm{Z}}{\Sigma_{t'}^i} {\mathbf{R}_{t'}^\mathbf{c}}^\top, \quad t' \coloneqq \argmin_{t' < t} |\prescript{}{\bm{Z}}{\Sigma_{t'}^i}|,
\end{equation}
which can be incrementally updated. $\mathbf{R}_t^\mathbf{c}$ is the rotational part of $\mathbf{P}_t^\mathbf{c}$. 

For each new frame, we perform motion prediction for each cluster using $\bm{v}_t^{\mathbf{q}}$. The predicted 3D landmark positions as well as its noise covariance matrix are re-projected back into the current frame using $\bm{\zeta}_t^i = \pi(\bm{p}_t^i + \bm{v}_t^\mathbf{q}), \Gamma_t^i = \mathbf{J}_\pi \Sigma_t^i \mathbf{J}_\pi^\top$.
The probability score of assigning the $k$-th observation to landmark $i$ becomes:
\begin{equation}
\label{equ:feat}
    p_i(k) \propto \left[ \lVert \bm{\zeta}_t^i - \bm{z}_t^k \rVert_{\Gamma_t^i}^2 < \gamma \right] \cdot s_{ik},
\end{equation}
where $[\cdot]$ is an indicator function, $s_{ik}$ is the descriptor similarity between landmark $i$ and observation $\bm{z}_t^k$ and $\gamma = 4.0$ in our experiments.
For each observation $k$, we choose its corresponding landmark $i$ with the highest assignment probability score: $k \rightarrow \argmax_{i} p_{i}(k) $ if possible.
In practice, Eq.~\ref{equ:feat} is only evaluated on a small neighbourhood of $\bm{z}_t^k$.

We further measure the uncertainty of the association $m \rightarrow \mathbf{q} \coloneqq \argmax_{\mathbf{q}'} p_{\mathbf{q}'}(m)$ by calculating the Shannon cross-entropy $\mathcal{E}_t^{\mathbf{q}}$ as:
\begin{equation}
\begin{aligned}
\mathcal{E}_t^{\mathbf{q}} &\coloneqq -\sum_m p_{\mathbf{q}}(m) \log p_{\mathbf{q}}(m), \\
p_{\mathbf{q}}(m) &\propto \sum_{\bm{\zeta}_t^k \in \bm{B}_t^m} (1 / |\Gamma_t^i|),
\end{aligned}
\end{equation}
where $p_{\mathbf{q}}(m)$ is the probability of assigning the $m$-th bounding box to cluster $\mathbf{q}$.
If $\mathcal{E}_t^{\mathbf{q}}$ is smaller than 1.0, we consider this as a successful high-level association, in which case we perform additional brute force low-level feature descriptor matching within the bounding box to find more feature correspondences.

\subsection{Heterogeneous CRF for Cluster Assignment}
\label{subsec:crf}

In this step, we determine the cluster assignment $\mathbf{q}^i$ of each landmark $i$ observed in the current frame. 
A conditional random field model combining semantic, spatial and motion information, which we call `heterogeneous CRF', is applied, minimizing the following energy:
\begin{equation}
    E(\{\mathbf{q}^i\}_i) \coloneqq \sum_i \psi_u(\mathbf{q}^i) + \alpha \sum_{i<j} \psi_p(\mathbf{q}^i, \mathbf{q}^j),
\end{equation}
which is a weighted sum ($\alpha>0$ being the balance factor) of unary energy $\psi_u$ and pairwise energy $\psi_p$ on a complete graph of all the observed landmarks.
The total number of classes for CRF is set to $M=N_1+N_2+1$, where $N_1$ is the number of \emph{live} clusters, $N_2$ is the number of unassociated bounding boxes in this frame and the trailing $1$ allows for an outlier class.
A cluster is considered \emph{live} if at least one of its landmarks is observed during the past $L$ frames.

\noindent\textbf{Unary Energy.}
The unary energy decides the probability of the observed landmark $i$ belonging to a specific cluster $\mathbf{q}^i$ and contains three sources of information:
\begin{equation}
\label{equ:crf}
    \psi_u(\mathbf{q}^i) \propto p_{\mathrm{2D}}(\mathbf{q}^i) \cdot p_{\mathrm{3D}}(\mathbf{q}^i) \cdot p_{\mathrm{mot}}(\mathbf{q}^i).
\end{equation}

The first multiplier $p_{\mathrm{2D}}$ incorporates information from the detected semantic bounding boxes. The probability should be large if the landmark lies within a bounding box.
Let $\mathbf{C}^i_t$ be the set of cluster indices corresponding to the bounding boxes where the observation of landmark $\bm{z}_t^i$ resides and $\eta$ be a constant for the detection confidence, then: 
\begin{equation}
    p_{\mathrm{2D}}(\mathbf{q}^i) \propto \begin{cases}
    \eta / |\mathbf{C}^i_t| & \mathbf{q}^i \in \mathbf{C}^i_t \\
    (1 - \eta) / (M - |\mathbf{C}^i_t|) & \mathbf{q}^i \notin \mathbf{C}^i_t
    \end{cases}.
\end{equation}

The second multiplier $p_{\mathrm{3D}}$ emphasizes the spatial affinity by assigning a high probability to the landmarks near the center of a cluster:
\begin{equation}
\label{equ:3dterm}
    p_{\mathrm{3D}}(\mathbf{q}^i) \propto \exp \left( -\lVert \bm{Z}_t^i - \bm{c}_{\mathbf{q}^i} \rVert^2_{\prescript{}{\bm{Z}}{\Sigma^i_t}}  / l_{\mathbf{q}^i}^2 \right ),
\end{equation}
where $\bm{c}_{\mathbf{q}^i}$ and $l_{\mathbf{q}^i}$ are the cluster center and dimension, respectively, determined by the center and the $30^{\mathrm{th}}$/$70^{\mathrm{th}}$ percentiles (found empirically) of the cluster landmark point cloud.

The third multiplier defines how the trajectories of cluster $\mathbf{q}^i$ over a set of timesteps $\mathcal{T}$ can explain the observation:
\begin{equation}
    p_{\mathrm{mot}}(\mathbf{q}^i) \propto \prod_{t'\in \mathcal{T}} \frac{\exp (-\lVert \bm{z}_{t'}^i - \pi(\mathbf{T}_{t'}^{\mathbf{c} \mathbf{q}^i} (\mathbf{P}_t^{\mathbf{q}^i})^{-1} \bm{p}_t^i) \rVert^2_{\prescript{}{\bm{z}}{\Sigma}} )}{\sqrt{|{}_{\bm{z}}\Sigma|}},
\end{equation}
which is a simple reprojection error w.r.t. the observations.
In our implementation we set $\mathcal{T} = \{ t-5, t \}$.
For the first 5 frames this term is not included in Eq.~\ref{equ:crf}.

The single 2D term only considers the 2D semantic detection, which possibly contains many outliers around the edge of the bounding box. 
By adding the 3D term, landmarks belonging to faraway background get pruned.
However, features close to the 3D boundary, \eg, on the ground nearby a moving vehicle, still have a high probability belonging to the cluster, whose confidence is further refined by the motion term.
Please refer to Sec.~\ref{subsec:ablation} for evaluations and visual comparisons on these three terms.

\noindent\textbf{Pairwise Energy.}
The pairwise energy is defined as:
\begin{equation}
    \psi_p(\mathbf{q}^i, \mathbf{q}^j) \coloneqq [\mathbf{q}^i \neq \mathbf{q}^j] \cdot \exp( - \lVert\bm{p}_t^i - \bm{p}_t^j \rVert^2),
\end{equation}
where the term inside the exponential operator is the distance between two landmarks $\bm{p}_t^i, \bm{p}_t^j$ in 3D space.
The pairwise energy can be viewed as a noise-aware Gaussian smoothing kernel to encourage spatial labeling continuity.

We use an efficient dense CRF inference method~\cite{krahenbuhl2011efficient} to solve for the energy minimization problem.
After successful inference, we perform Kuhn-Munkres algorithm
to match current CRF clustering results with previous cluster assignments. New clusters are created if no proper cluster assignment is found for an inferred label.
We then update the weight $w^i$ for each landmark according to a strategy introduced in \cite{tateno2015real} and change its cluster assignment if necessary:
When the newly assigned cluster is the same as the landmark's previous cluster, we increase the weight $w^i$ by 1, otherwise the weight is decreased by 1.
When $w^i$ is decreased to 0, a change in cluster assignment is triggered to accept the currently assigned cluster.

\subsection{Sliding-Window State Estimation}
\label{subsec:slam}

\noindent\textbf{Double-Track Frame Management.}
Keyframe-based SLAM systems like ORB-SLAM2~\cite{mur2017orb} select keyframes by the spatial distance between frames and the number of commonly visible features among frames.
For ClusterVO where the trajectory of each cluster is incorporated into the state estimation process, the aforementioned strategy for keyframe selection is not enough to capture the relatively fast-moving clusters.

\begin{figure}
    \centering
    \includegraphics[width=\linewidth]{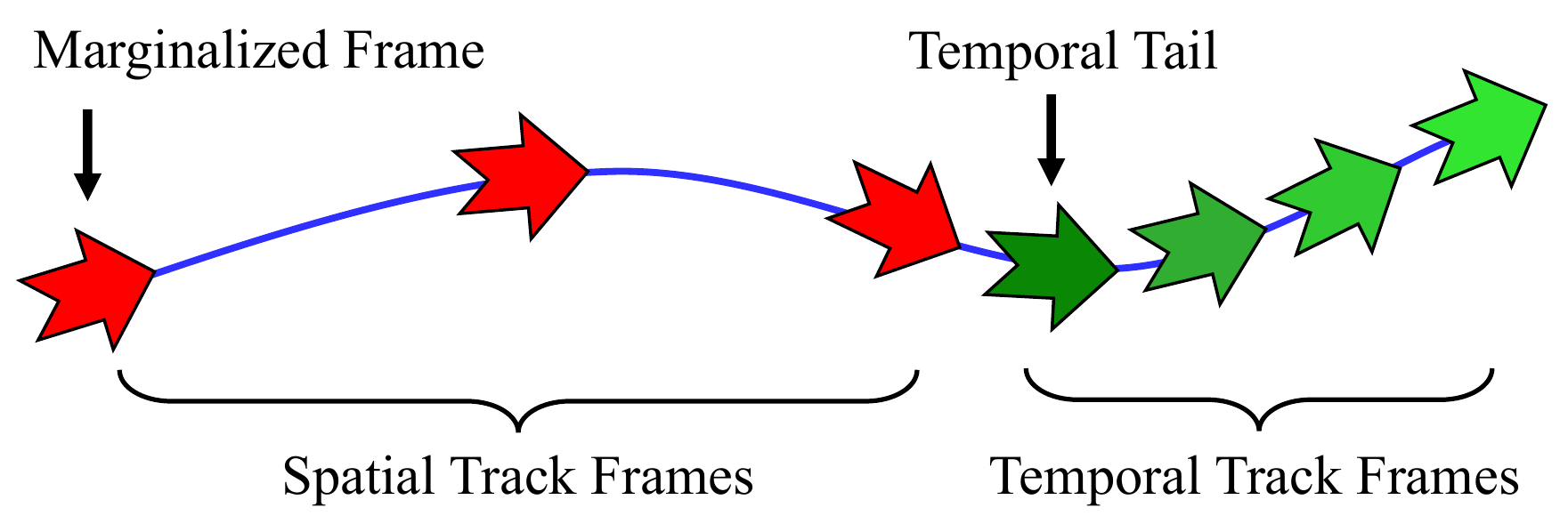}
    \caption{\textbf{Frame Management in ClusterVO.} Frames maintained by the system consist of spatial track (red) and temporal track (green). When a new frame comes, the oldest frame in the temporal track (Temporal Tail) will either be discarded or promoted into the spatial track. The last spatial frame is to be marginalized if the total number of spatial frames exceeds a given threshold.}
    \label{fig:mgmt}
\end{figure}

Instead of the chunk strategy proposed in ClusterSLAM~\cite{huang2019clusterslam}, we employ a sliding window optimization scheme in accordance with a novel \emph{double-track} frame management design (Figure~\ref{fig:mgmt}).
The frames maintained and optimized by the system are divided into two sequential tracks: a temporal track $\mathcal{T}_t$ and a spatial track $\mathcal{T}_s$. 
$\mathcal{T}_t$ contains the most recent input frames.  
Whenever a new frame comes, the oldest frame in $\mathcal{T}_t$ will be moved out. If this frame is spatially far away enough from the first frame in $\mathcal{T}_s$ or the number of commonly visible landmarks is sufficiently small, this frame will be appended to the tail of $\mathcal{T}_s$, otherwise it will be discarded.
This design has several advantages.
First, frames in the temporal track record all recent observations and hence allow for enough observations to track a fast-moving cluster.
Second, previous wrongly clustered landmarks can be later corrected and re-optimization based on new assignments is made possible.
Third, features detected in the spatial track help create enough parallax for accurate landmark triangulation and state estimation.

\noindent\textbf{For static scene $\mathbf{q}=0$ and camera pose},
the energy function for optimization is a standard Bundle Adjustment ~\cite{triggs1999bundle} augmented with an additional marginalization term:
\begin{equation}
    \begin{aligned}
    \label{equ:static_optim}
    \mathbf{E}(\{\bm{x}_{t}^\mathbf{c},\bm{x}_{t}^\mathbf{L}\}_{t\in \mathcal{T}_a }) \coloneqq & \sum_{i \in \mathcal{I}_0,t\in \mathcal{T}_a} \rho( \lVert \bm{z}_t^i - \pi ( (\mathbf{P}_t^\mathrm{c})^{-1} \bm{p}_t^i ) \rVert^2_{\prescript{}{\bm{z}}{\Sigma}} ) \\ 
    +&\sum_{t\in \mathcal{T}_a} \lVert \delta {\bm{x}_t^\mathbf{c}} - \mathbf{H}^{-1}\bm{\beta} \rVert^2_{\mathbf{H}},
\end{aligned}
\end{equation}
where $\mathcal{T}_a \coloneqq \mathcal{T}_s \cup \mathcal{T}_t$, $\mathcal{I}_\mathbf{q} = \{i | \mathbf{q}^i=\mathbf{q}\}$ indicates all landmarks belonging to cluster $\mathbf{q}$ and $\rho(\cdot)$ is robust Huber M-estimator. 
As the static scene involves a large number of variables and simply dropping these variables out of the sliding window will cause information loss, leading to possible drifts, we \emph{marginalize} some variables which would otherwise be removed and summarize the influence to the system with the marginalization term in Eq.~\ref{equ:static_optim}.
Marginalization is only performed when a frame is discarded from the spatial track $\mathcal{T}_s$.
To restrict dense fill-in of landmark blocks in the information matrix, the observations from the frame to be removed will be either deleted if the corresponding landmark is observed by the newest frame or marginalized otherwise. This marginalization strategy only adds dense Hessian block onto the frames instead of landmarks, making the system still solvable in real-time.

More specifically, in the marginalization term, $\delta \bm{x}$ is the state change relative to the critical state $\bm{x}^*$ captured when marginalization happens.
For the computation of $\mathbf{H}$ and $\bm{\beta}$, we employ the standard Schur Complement: $\mathbf{H} = \mathbf{\Lambda}_{aa} - \mathbf{\Lambda}_{ab} \mathbf{\Lambda}_{bb}^{-1} \mathbf{\Lambda}_{ba}, \bm{\beta} = \bm{b}_a  - \mathbf{\Lambda}_{ab} \mathbf{\Lambda}_{bb}^{-1} \bm{b}_b$, where $\mathbf{\Lambda}_{(\cdot)}$ and $\bm{b}_{(\cdot)}$ are components of the system information matrix $\mathbf{\Lambda}$ and information vector $\bm{b}$ extracted by linearizing around $\bm{x}^*$: 
\begin{equation}
    \bm{\Lambda} = \begin{bmatrix}
    \mathbf{\Lambda}_{aa} & \mathbf{\Lambda}_{ab} \\
    \mathbf{\Lambda}_{ba} & \mathbf{\Lambda}_{bb} \\
    \end{bmatrix}, \quad
    \bm{b} = \begin{bmatrix}
    \bm{b}_{a} \\ \bm{b}_{b}
    \end{bmatrix}.
\end{equation}

\noindent\textbf{For dynamic clusters $\mathbf{q}\neq0$}, 
the motions are modeled using a white-noise-on-acceleration prior~\cite{barfoot2017state}, 
which can be written in the following form in continuous time $t,t'\in\mathbb{R}$:
\begin{equation}
\label{equ:gp}
    \ddot{\bm{t}^\mathbf{q}}(t) \sim \mathcal{GP}(\bm{0}, \bm{Q}\delta(t-t')),
\end{equation}
where $\bm{t}^\mathbf{q}$ is the translational part of the continuous cluster pose $\mathbf{P}^\mathbf{q}$ (hence $\ddot{\bm{t}^\mathbf{q}}$ is the cluster acceleration),  $\mathcal{GP}$ stands for the Gaussian Process, and $\bm{Q}$ denotes its power spectral matrix. We define the energy function for optimizing the $\mathbf{q}$-th cluster trajectories and its corresponding landmark positions as follows:
\begin{equation}
\begin{aligned}
\label{equ:cluster_optim}
    \mathbf{E}(\{ \bm{x}_t^\mathbf{q}, \bm{x}_t^\mathbf{L} \}_{t \in \mathcal{T}_t}) \coloneqq & \sum_{t,t^{+} \in \mathcal{T}_t} \left \lVert \begin{bmatrix}\bm{t}_{t^+}^i\\\bm{v}_{t^+}^i\end{bmatrix} - \mathbf{A} \begin{bmatrix}\bm{t}_{t}^i\\\bm{v}_{t}^i\end{bmatrix} \right \rVert^2_{\hat{\bm{Q}}} \\ + \sum_{i\in \mathcal{I}_\mathbf{q},t\in\mathcal{T}_t} & \rho( \lVert \bm{z}_t^i - \pi ( \mathbf{T}_t^{\mathbf{c}\mathbf{q}^i} (\mathbf{P}_t^\mathbf{c})^{-1} \bm{p}_t^i ) \rVert^2_{\prescript{}{\bm{z}}{\Sigma}} ),
\end{aligned}
\end{equation}
in which
\begin{equation}
    \mathbf{A} \coloneqq \begin{bmatrix} \mathbf{I} & \Delta t \mathbf{I} \\ \mathbf{0} & \mathbf{I} \end{bmatrix},  \hat{\bm{Q}}^{-1} \coloneqq \begin{bmatrix} 12 / \Delta t^{3}  & -6 / \Delta t^{2} \\ -6 / \Delta t^{2} & 4 / \Delta t \end{bmatrix} \otimes \bm{Q}^{-1},
\end{equation}
where $\otimes$ is the Kronecker product and $\Delta t = t^+ - t$, $t^+$ being the next adjacent timestamp of frame $t$.
Eq.~\ref{equ:cluster_optim} is the sum of motion prior term and reprojection term. The motion prior term is obtained by querying the random process model of Eq.~\ref{equ:gp}, which intuitively penalizes the change in velocity over time and smooths cluster motion trajectory which would otherwise be noisy due to fewer features on clusters than static scenes.
Note that different from the energy term for the static scene which optimizes over both $\mathcal{T}_s$ and $\mathcal{T}_t$, for dynamic clusters only $\mathcal{T}_t$ is considered.

During the optimization of cluster state, the camera state $\bm{x}_t^\mathbf{c}$ stays unchanged. The optimization process for each cluster can be easily paralleled because their states are mutually independent (in practice the system speed is 8.5Hz \& 7.8Hz for 2 \& 6 clusters, resp.).

\section{Experiments}

\subsection{Datasets and Parameter Setup}

The effectiveness and general applicability of ClusterVO system is mainly demonstrated in two scenarios: indoor scenes with moving objects and autonomous driving with moving vehicles.

For indoor scenes, we employ the stereo Oxford Multimotion dataset (OMD)~\cite{judd2019oxford} for evaluation.
This dataset is specially designed for indoor simultaneous camera localization and rigid body motion estimation, with the ground-truth trajectories recovered using a motion capture system.
Evaluations and comparisons are performed on two sequences: \texttt{swinging\_4\_unconstrained} (S4, 500 frames, with four moving bodies: S4-C1, S4-C2, S4-C3, S4-C4) and \texttt{occlusion\_2\_unconstrained} (O2, 300 frames, with two moving bodies: O2-Tower and O2-Block), because these are the only sequences with baseline results reported in sequential works from Judd~\etal~\cite{judd2018multimotion,judd2019occlusion} named `MVO'.

For autonomous driving cases, we employ the challenging KITTI dataset~\cite{geiger2012we} for demonstration.
As most of the sequences in the odometry benchmark have low dynamics and comparisons on these data can hardly lead to sensible improvements over other SLAM solutions (\eg ORB-SLAM), similar to Li~\etal~\cite{li2018stereo}, we demonstrate the strength of our method in selected sequences from the raw dataset as well as the full 21 tracking training sequences with many moving cars.
The ground-truth camera ego-motion is obtained from the OxTS packets (combining GNSS and inertial navigation) provided by the dataset.

The CRF weight is set to $\alpha=5.0$ and the 2D unary energy constant $\eta = 0.95$.
The power spectral matrix $\bm{Q} = 0.01\mathbf{I}$ for the motion prior.
The maximum sizes of the double-track are set to $|\mathcal{T}_s| = 5$ and $|\mathcal{T}_t| = 15$.
The threshold for determining whether the cluster is still live is set to $L=|\mathcal{T}_t|$.
All of the experiments are conducted on an Intel Core i7-8700K, 32GB RAM desktop computer with an Nvidia GTX 1080 GPU.

\subsection{Indoor Scene Evaluations}

We follow the same evaluation protocol as in~\cite{judd2018multimotion}, by computing the maximum drift (deviation from ground-truth pose) across the whole sequence in translation and rotation (represented in three Euler angles, namely roll, yaw and pitch) for camera ego-motion as well as for all moving cluster trajectories.
As our method does not define a canonical frame for detected clusters, we need to register the pose recovered by our method with the ground-truth trajectory.
To this end, we multiply our recovered pose with a rigid transformation $\mathbf{T}_\mathrm{r}$ which minimizes the sum of the difference between $\mathbf{P}_t^\mathbf{q} \mathbf{T}_\mathrm{r}$ and the ground-truth pose for all $t$.
This is based on the assumption that the local coordinates of the recovered landmarks can be registered with the positions of ground-truth landmarks using this rigid transformation.

{For the semantic bounding box extraction}, the YOLOv3 network~\cite{redmon2017yolo9000} is {re-trained} to detect an additional class named `block' representing the swinging or rotating blocks in the dataset. The detections used for training are labeled using a combined approach with human annotations and a median flow tracker on the rest frames from S4 and O2.

\begin{figure}[t]
    \centering
    \includegraphics[width=\linewidth]{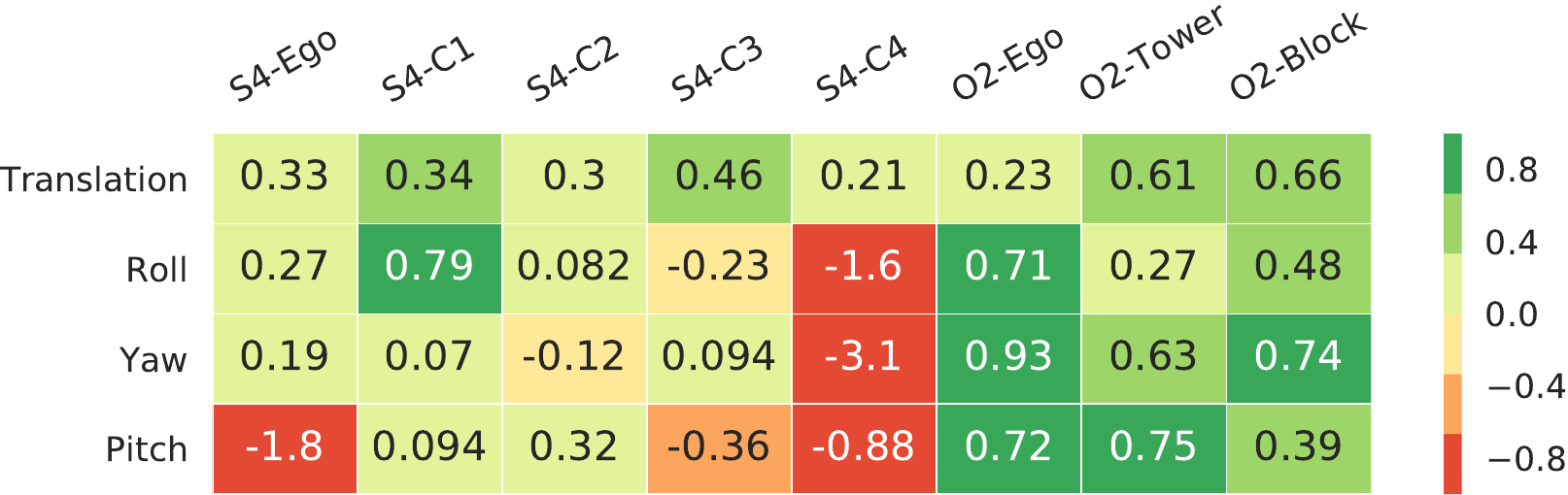}
    \caption{Performance comparison with MVO on S4 and O2 sequence in Oxford Multimotion~\cite{judd2019oxford} dataset. The numbers in the heatmap show the ratio of decrease in error using ClusterVO for different trajectories and measurements.}
    \label{fig:omd-eval}
\end{figure}

\begin{figure}[t]
    \centering
    \includegraphics[width=1.0\linewidth]{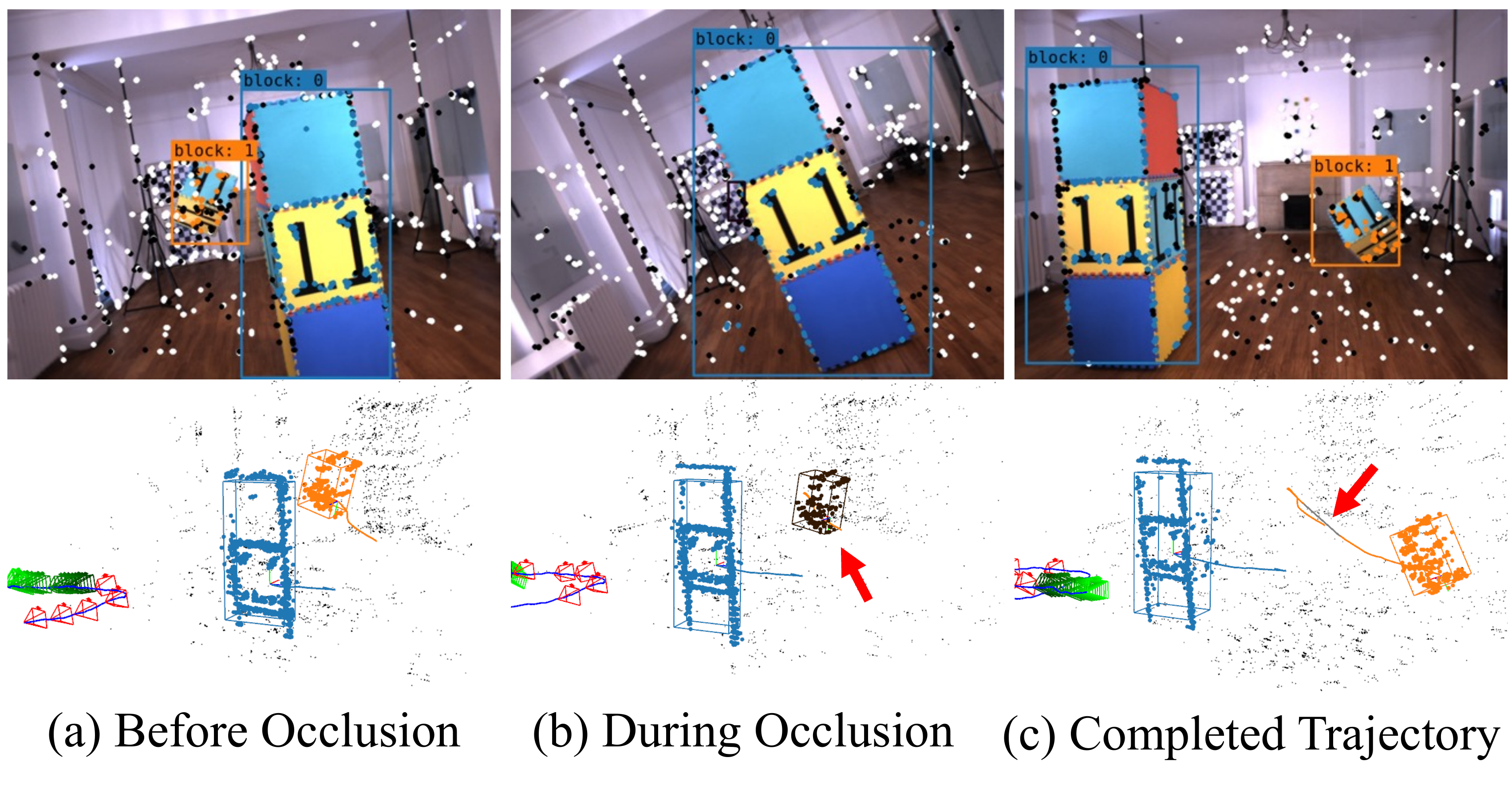}
    \caption{Qualitative results in OMD Sequence O2. The three subfigures demonstrate an occlusion handling process by ClusterVO.}
    \label{fig:o2-result}
\end{figure}

\begin{figure}
    \centering
    \includegraphics[width=0.95\linewidth]{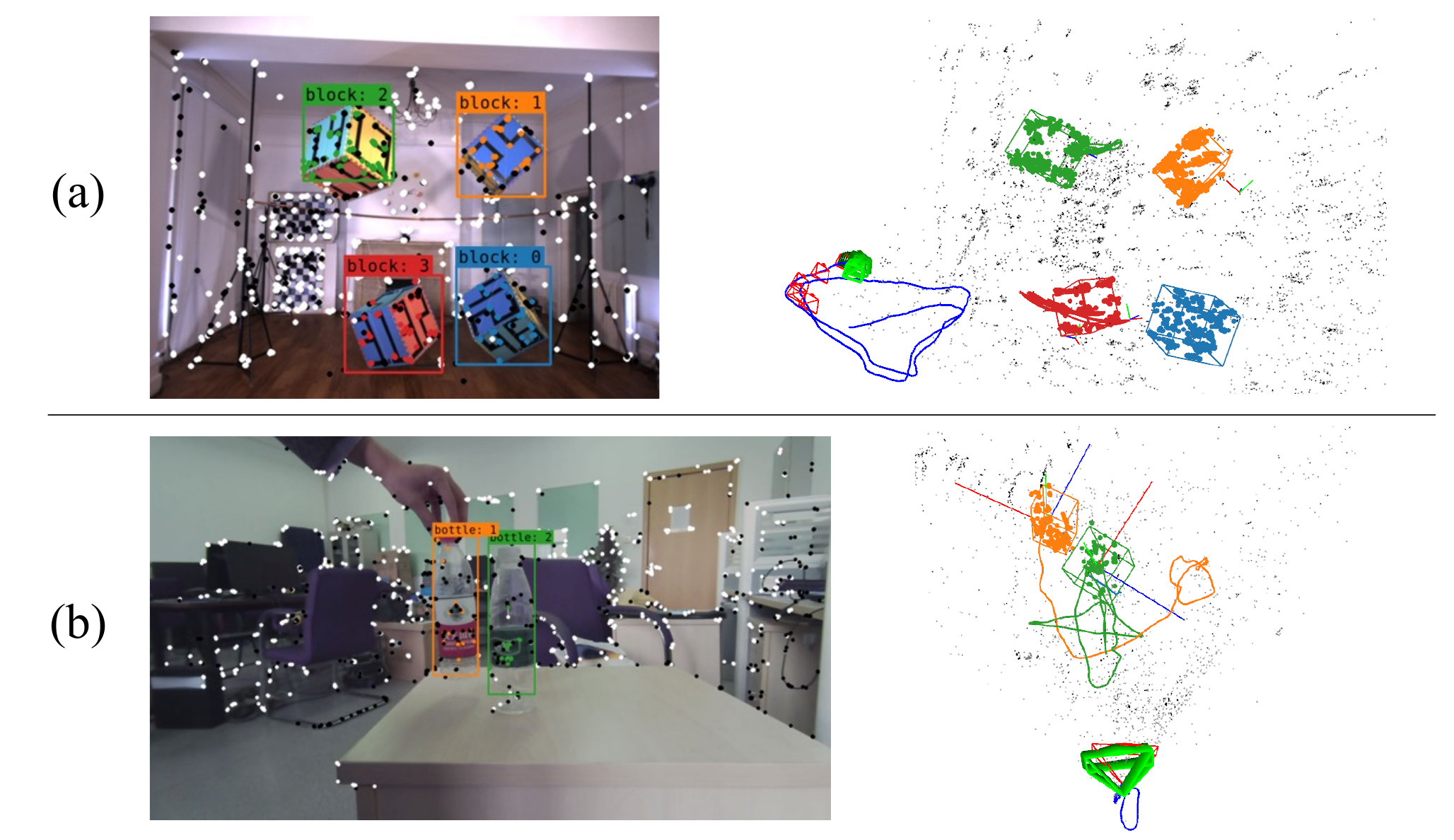}
    \caption{Other indoor qualitative results. (a) OMD Sequence S4; (b) A laboratory scene where two bottles are reordered.}
    \label{fig:indoor-result}
\end{figure}

\begin{table*}
\centering
\caption{Camera ego-motion comparison with state-of-the-art systems on KITTI raw dataset. The unit of ATE and T.RPE is meters and the unit for R.RPE is radians.}
\label{tbl:odometry}
\footnotesize
\begin{tabular}{c|ccc|ccc|c|ccc|ccc} 
\hline
\multirow{2}{*}{Sequence} & \multicolumn{3}{c|}{ORB-SLAM2~\cite{mur2017orb}}                & \multicolumn{3}{c|}{DynSLAM~\cite{barsan2018robust}}  & Li \etal~\cite{li2018stereo}  & \multicolumn{3}{c|}{ClusterSLAM~\cite{huang2019clusterslam}}  & \multicolumn{3}{c}{ClusterVO}                  \\
                          & ATE           & R.RPE         & T.RPE         & ATE   & R.RPE & T.RPE          & ATE           & ATE           & R.RPE         & T.RPE      & ATE           & R.RPE         & T.RPE          \\ 
\hline
0926-0009                 & 0.91          & \textbf{0.01} & \textbf{1.89} & 7.51  & 0.06  & 2.17           & 1.14   & 0.92 & 0.03 & 2.34       & \textbf{0.79} & 0.03          & 2.98           \\
0926-0013                 & 0.30          & \textbf{0.01} & \textbf{0.94} & 1.97  & 0.04  & 1.41           & 0.35   & 2.12 & 0.07 & 5.50       & \textbf{0.26} & \textbf{0.01} & 1.16           \\
0926-0014                 & 0.56          & \textbf{0.01} & 1.15      & 5.98  & 0.09  & 2.73           & 0.51   & 0.81 & 0.03 & 2.24       & \textbf{0.48} & \textbf{0.01} & \textbf{1.04}  \\
0926-0051                 & \textbf{0.37} & \textbf{0.00} & \textbf{1.10} & 10.95 & 0.10  & 1.65           & 0.76   & 1.19 & 0.03 & 1.44      & 0.81          & 0.02          & 2.74           \\
0926-0101                 & 3.42          & 0.03          & 14.27         & 10.24 & 0.13  & \textbf{12.29} & 5.30   & 4.02 & \textbf{0.02} & 12.43       & \textbf{3.18} & \textbf{0.02} & 12.78          \\
0929-0004                 & 0.44          & \textbf{0.01} & \textbf{1.22} & 2.59  & 0.02  & 2.03           & \textbf{0.40} & 1.12 & 0.02 & 2.78 & \textbf{0.40} & 0.02          & 1.77           \\
1003-0047                 & 18.87         & 0.05          & 28.32         & 9.31  & 0.05  & 6.58           & \textbf{1.03} & 10.21 & 0.06 & 8.94 & 4.79          & 0.05          & \textbf{6.54}  \\ 
\hline
\end{tabular}
\end{table*}

Figure~\ref{fig:omd-eval} shows the ratio of decrease in the drift compared with the baseline MVO~\cite{judd2018multimotion,judd2019occlusion}.
More than half of the trajectory estimation results improve by over 25\%, leading to accurate camera ego-motion and cluster motion recoveries. 
Two main advantages of ClusterVO over {MVO} have made the improvement possible: 
First, the pipeline in {MVO} requires a stable tracking of features in each input batch of $\sim$50 frames and this keeps only a small subset of landmarks where the influence of noise becomes more dominating, while ClusterVO maintains consistent landmarks for each individual cluster and associates both low-level and high-level information to maximize the utility of historical information.
Second, if the motion in a local window is small, {the geometric-based method} will tend to misclassify dynamic landmarks and degrade the recovered pose results; ClusterVO, however, leverages additional semantic and spatial information to achieve more accurate and meaningful classification and estimation.

Meanwhile, the robust association strategy and double-track frame management design allow ClusterVO to continuously track cluster motion even it is temporarily occluded. This feature is demonstrated in figure \ref{fig:o2-result} on the O2 sequence where the block is occluded by the tower for $\sim$10 frames.
The cluster's motion is predicted during the occlusion and finally the prediction is probabilistically associated with the re-detected semantic bounding box of the block.
The state estimation module {is then relaunched} to recover the motion using the information both before and after the occlusion.

Figure~\ref{fig:indoor-result}(a) shows qualitative results on the S4 sequence and in Figure~\ref{fig:indoor-result}(b) another result from a practical indoor laboratorial scene with two moving bottles recorded using a Mynteye stereo camera is shown.

\subsection{KITTI Driving Evaluations}

Similar to Li \etal~\cite{li2018stereo}, we divide the quantitative evaluation into ego-motion comparisons and 3D object detection comparisons.
Our results are compared to state-of-the-art systems including ORB-SLAM2~\cite{mur2017orb}, DynSLAM~\cite{barsan2018robust}, Li~\etal~\cite{li2018stereo} and ClusterSLAM~\cite{huang2019clusterslam} using the TUM metrics~\cite{sturm2012benchmark}.
These metrics evaluate ATE, R.RPE and T.RPE, which are short for the Root Mean Square Error (RMSE) of the Absolute Trajectory Error, the Rotational and Translational Relative Pose Error, respectively.

As shown in Table~\ref{tbl:odometry}, for most of the sequences we achieve the best results in terms of ATE, meaning that our method can maintain globally correct camera trajectories in challenging scenes (\eg 1003-0047) where even ORB-SLAM2 fails due to its static scene assumption.
Although DynSLAM maintains a dense mapping of both the static scenes and dynamic objects, the underlying sparse scene flow estimation is based on a frame-to-frame visual odometry \emph{libviso}~\cite{geiger2011stereoscan}, which will inherently lead to remarkable drift over long travel distances.
The batch Multibody SfM formulation of Li~\etal results in a highly nonlinear factor graph optimization problem whose solution is not trivial. 
ClusterSLAM~\cite{huang2019clusterslam} requires associated landmarks and the inaccurate feature tracking frontend affects the localization performance even if the states are solved via full optimization.
In contrast, our ClusterVO achieves comparable or even better results than all previous methods due to the fusing of multiple sources of information and the robust sliding-window optimization.

The cluster trajectories are evaluated in 3D object detection benchmark in KITTI tracking dataset.
We compute the Average Precision (AP) of the `car' class in both bird view ($\mathrm{AP_{bv}}$) and 3D view ($\mathrm{AP_{3D}}$).
Our detected 3D box center is $\bm{c}_\mathrm{q}$ (in Eq.~\ref{equ:3dterm}) and the dimension is taken as the average car size. The box orientation is initialized to be vertical to the camera and tracked over time later on.
The detection is counted as a true positive if the Intersection over Union (IoU) score with an associated ground-truth detection is larger than 0.25.
All ground-truth 3D detections are divided into three categories (Easy, Moderate and Hard) based on the height of 2D reprojected bounding box and the occlusion/truncation level.

We compare the performance of our method with the state-of-the-art 3D object detection solution from Chen \etal~\cite{chen20173d} and DynSLAM~\cite{barsan2018robust}. 
The evaluation is performed in camera coordinate system so the inaccuracies in ego-motion estimations are eliminated.

\begin{table}[t]
\centering
\setlength{\tabcolsep}{0.8pt}
\caption{3D object detection comparison on KITTI dataset.}
\label{tbl:tracking}
\small
\begin{tabular}{c|ccc|ccc|c} 
\hline
\multicolumn{1}{l|}{}                  & \multicolumn{3}{c|}{$\mathrm{AP_{bv}}$ } & \multicolumn{3}{c|}{$\mathrm{AP_{3D}}$ } & \multicolumn{1}{l}{\multirow{2}{*}{\begin{tabular}[c]{@{}c@{}}Time\\(ms)\end{tabular}}}  \\
\multicolumn{1}{l|}{}                  & Easy  & Moderate & Hard                  & Easy  & Moderate & Hard                  & \multicolumn{1}{l}{}                       \\ 
\hline
Chen~\etal~\cite{chen20173d} & \textbf{81.34} & \textbf{70.70}    & \textbf{66.32}                 & \textbf{80.62} & \textbf{70.01}    & \textbf{65.76}                 & 1200                                          \\
DynSLAM~\cite{barsan2018robust}               & 71.83 & 47.16    & 40.30                 & 64.51 & 43.70    & 37.66                 & 500                                          \\ 
\hline
ClusterVO                              & 74.65 & 49.65    & 42.65                 & 55.85 & 38.93    & 33.55                 & \textbf{125}                                          \\
\hline
\end{tabular}
\end{table}

The methods of Chen~\etal and DynSLAM are similar in that they both perform a dense stereo matching (\eg \cite{yamaguchi2014efficient}) to precompute the 3D structure. 
While DynSLAM crops the depth map using 2D detections to generate spatial detections, Chen~\etal generates and scores object proposals directly in 3D space incorporating many scene priors in autonomous driving scenarios including the ground plane and car dimension prior.
These priors are justified to be critical comparing the results in Table~\ref{tbl:tracking}: DynSLAM witnesses a sharp decrease in both Moderate and Hard categories which contain faraway cars and small 2D detection bounding boxes.

\begin{figure}
    \centering
    \includegraphics[width=1.0\linewidth]{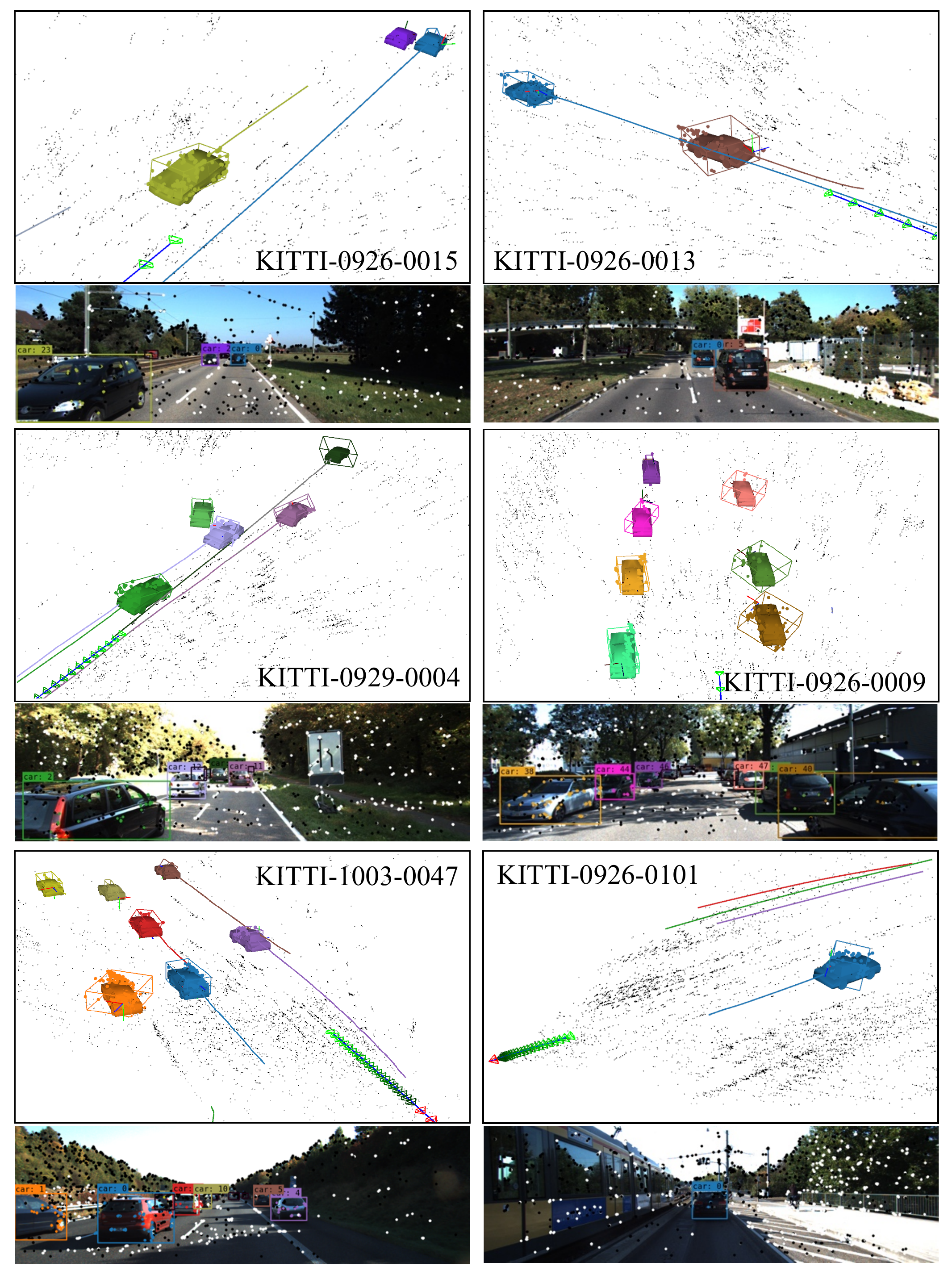}
    \caption{Qualitative results on KITTI raw dataset. The image below each sequence shows the input image and detections of the most recent frame.}
    \label{fig:kitti-result}
    \vspace{-1em}
\end{figure}

In the case of ClusterVO, which is designed to be general-purpose, the natural uncertainty of stereo triangulation becomes larger when the landmark becomes distant from the camera without object size priors. Also, we do not detect the canonical direction (\ie, the front of the car) of the cluster if its motion is small, so the orientation can be imprecise as well.
This explains the gap in detecting hard examples between ours and a specialized system like \cite{chen20173d}.
Compared to DynSLAM, the average precision improves because ClusterVO is able to track the moving object over time consistently and predicts their motions even if the 2D detection network misses some targets.
Additionally, we emphasize the high efficiency of ClusterVO system by comparing the time cost in Table \ref{tbl:comp} while the work of Chen~\etal requires 1.2 seconds for each stereo input pair.
Some qualitative results of KITTI raw dataset are shown in Figure~\ref{fig:kitti-result}. 

\subsection{Ablation study}
\label{subsec:ablation}

We test the importance of each probabilistic term in our Heterogeneous CRF formulation (Eq. \ref{equ:crf}) using synthesized motion dataset rendered from SUNCG~\cite{song2017semantic}.
Following the same stereo camera parameter as in \cite{huang2019clusterslam}, we generate 4 indoor sequences with moving chairs and balls, and compare the accuracies of ego motion and cluster motions in Table~\ref{tbl:ablation}.

\begin{table}[t]
\caption{Ablation comparisons on SUNCG dataset in terms of ego-motion and cluster trajectories.}
\label{tbl:ablation}
\footnotesize
\centering
\begin{tabular}{l|cccc}
\hline
                 & \multicolumn{2}{c}{Ego Motion$^\star$} & \multicolumn{2}{c}{Cluster Motion} \\
                 & ATE         & R./T.RPE              & ATE           & R./T.RPE                \\ \hline
ORB-SLAM2~\cite{mur2017orb}        &   \textbf{0.35}    &   0.14/\textbf{0.59}               &    -    &      -       \\
DynSLAM~\cite{barsan2018robust}         &  54.07     &   11.07/49.24       &    0.26    & 1.23/0.59          \\
ClusterSLAM~\cite{huang2019clusterslam} &   1.34    &   0.41/1.89       &   0.17   &   0.34/\textbf{0.30}     \\ \hline
ClusterVO 2D     & 0.62        & 0.19/0.95        & 0.24          & \textbf{0.31}/0.53          \\
ClusterVO 2D+3D  &  0.52    &    \textbf{0.11}/0.87     &     0.15    &    0.50/0.53    \\
ClusterVO Full   &   0.61   &    0.19/0.91     &     \textbf{0.13}   &     0.37/0.36      \\ \hline

\end{tabular} \\
\vspace{-1.2em}
\flushleft\footnotesize{
$^\star$ Values are multiplied by 100.}
\end{table}

\begin{figure}
    \centering
    \includegraphics[width=1.0\linewidth]{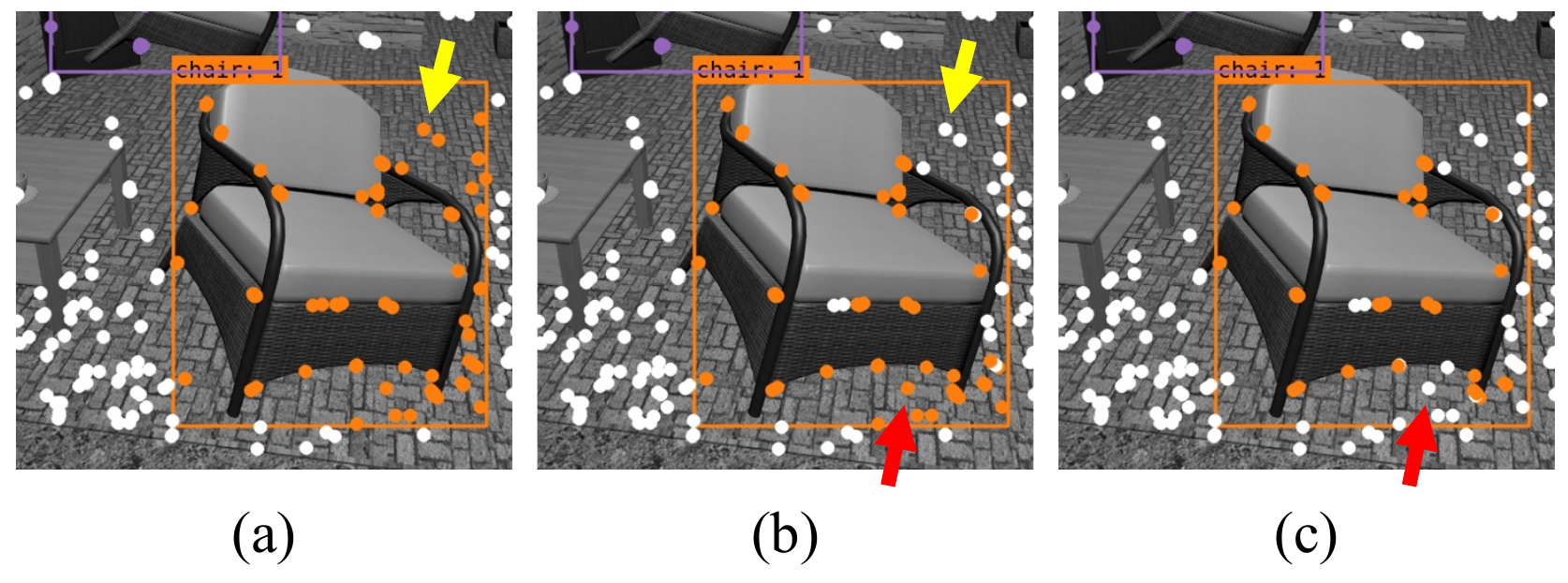}
    \caption{Unary term visualizations on one indoor sequence from SUNCG dataset. (a) ClusterVO 2D; (b) ClusterVO 2D+3D; (c) ClusterVO Full.}
    \label{fig:ablation}
    \vspace{-1em}
\end{figure}

By gradually adding different terms of Eq.~\ref{equ:crf} into the system, our performance on estimating cluster motions improves especially in terms of absolute trajectory error (decreases by 45.8\% compared to 2D only CRF) while the accuracy of ego motion is not affected.
This is due to the more accurate moving object clustering combining both geometric and semantic cues.
It should be noted that our results are even comparable to the most recent ClusterSLAM~\cite{huang2019clusterslam}, a backend method with full batched Bundle Adjustment optimization: 
This shows that incorporating semantic information into the motion detection problem helps effectively regularize the solution and achieves more consistent trajectory estimation.
Figure~\ref{fig:ablation} visualizes this effect further by computing the classification result based only on the unary term $\psi_u$.
Some mis-classified landmarks are successfully filtered out by incorporating more information.

\section{Conclusion}

In this paper we present ClusterVO, a general-purpose fast stereo visual odometry for simultaneous moving rigid body clustering and motion estimation.
Comparable results to state-of-the-art solutions on both camera ego-motion and dynamic objects pose estimation demonstrate the effectiveness of our system.
In the future, one direction would be to incorporate specific scene priors as pluggable components to improve ClusterVO performance on specialized applications (\eg autonomous driving); another direction is to fuse information from multiple sensors to further improve localization accuracy.


\vspace{0.5em}
\noindent\textbf{Acknowledgements.}
We thank anonymous reviewers for the valuable discussions.
This work was supported by the Natural Science Foundation of China (Project Number 61521002, 61902210), the Joint NSFC-DFG Research Program (Project Number 61761136018) and Research Grant of Tsinghua-Tencent Joint Laboratory for Internet Innovation Technology.

{\small
\bibliographystyle{ieee_fullname}
\bibliography{main}
}

\end{document}